# Ingredient-Level Food Image Segmentation for Nutrition Awareness

Jonesh Shrestha
Jarvis College of Computing and Digital Media
DePaul University
Chicago Illinois United States
jshresth@depaul.edu

## ABSTRACT

Food images often contain several visible ingredients, so assigning one dish label to an entire image hides important visual structure. This work studies ingredient-level semantic segmentation on FoodSeg103, where the model predicts an ingredient class for each pixel. Two SegFormer variants were fine-tuned and evaluated under a controlled setup: SegFormer-B0 as the smaller baseline model and SegFormer-B1 as the larger final model. Both models use ImageNet-pretrained MiT backbones with newly initialized 104-class output layers. On the held-out FoodSeg103 test split of 2,135 images, B0 achieved 0.7709 pixel accuracy and 0.2521 mean IoU, while B1 achieved 0.7929 pixel accuracy and 0.3204 mean IoU. B1 improved every saved test metric, including a +0.0683 absolute gain in mean IoU. The system also converts predicted masks into visible ingredient-area percentages, giving a simple visual composition summary of the predicted meal. This summary can serve as a first-pass nutrition-awareness cue by providing a visual alternative to detailed food tracking similar to plate-based meal guidance, but it is not a direct estimate of calories, macronutrients, food mass, volume, density, or true portion size.

## KEYWORDS

## 1 Introduction

Food recognition is often treated as image classification: given a picture, a model predicts one dish label. That is useful for retrieval or broad recognition, but it is limited when a meal contains several visible ingredients. A single label such as "salad" or "meal" does not show where the rice, vegetables, meat, sauce, fruit, or background appear in the image. For nutrition awareness, a more useful first step is to separate the visible ingredients at the pixel level.

This study focuses on ingredient-level semantic segmentation. Semantic segmentation means assigning a class label to every pixel in an image. In this experiment, the labels come from FoodSeg103, so the output is a mask that separates visible food regions into ingredient classes. From that mask, the system also computes a visible-area summary: among the pixels predicted as food, what percentage is assigned to each ingredient?

The scope is intentionally narrow, but still meaningful. Detailed calorie and macronutrient tracking can be time-consuming for everyday use, which is one reason visual plate-based guidance remains useful. The Healthy Eating Plate, for example, uses approximate visual proportions to help people reason about meal balance without requiring exact calorie or serving calculations [6]. This work follows a similar idea from a computer-vision perspective: predicted masks are used to summarize which visible ingredients occupy more or less image area. However, a single RGB image does not provide depth, density, mass, or real volume. Therefore, the system should be interpreted as an ingredient-segmentation and visible-composition proof of concept rather than a complete nutrition-estimation system. The main contributions are:

1. A reproducible FoodSeg103 fine-tuning pipeline for ingredient-level segmentation using SegFormer.
2. A controlled comparison between SegFormer-B0 and SegFormer-B1, where model capacity changes while the data split, training setup, and evaluation code stay fixed.

3. Held-out test evaluation at original mask resolution, including pixel accuracy, mean IoU, foreground mean IoU, mean class accuracy, and per-class behavior.
4. A visible ingredient-area module that turns predicted masks into an interpretable 2D composition summary for nutrition awareness with some limitations.

The research questions are: (1) Can SegFormer be fine-tuned for ingredient-level segmentation on FoodSeg103? (2) Does the larger SegFormer-B1 model improve over SegFormer-B0 under the same experimental setup? (3) Can predicted masks support a useful visible-area summary?

## 2 Literature Review

Fully Convolutional Networks (FCNs) helped establish modern deep semantic segmentation by converting classification networks into end-to-end dense predictors by predicting a class at many spatial locations, ideally every pixel, instead of producing only one label for the whole image [3]. Later architectures improved segmentation with stronger multi-scale context and boundary refinement. DeepLabV3+ is a useful example because it combines atrous spatial pyramid pooling for multi-scale context with a decoder designed to refine object boundaries [5]. These works provide background for dense prediction, but they are not implemented as controlled baselines in this study.

FoodSeg103 is the main dataset for this study [1]. It was introduced to support fine-grained food image segmentation with ingredient-level pixel annotations. This is essential because food images are not simple object images: ingredients can overlap, appear in mixed dishes, look similar to other ingredients, or vary in appearance depending on preparation. Recent FoodSeg103 work comparing transformer-based and convolutional approaches also emphasizes these same difficulties, including overlap, inter-class similarity, intra-class variation, and class imbalance [4].

SegFormer is the model family used in this experiment [2]. It combines a hierarchical Mix Transformer encoder with a lightweight MLP decoder. In simpler terms, the encoder extracts visual features at multiple scales, and the decoder turns those features into a pixel-level class prediction map. This design is appropriate for ingredient segmentation as the model needs both local detail for boundaries and broader context for recognizing ingredients in mixed dishes.

This paper reports two SegFormer variants. SegFormer-B0 / MiT-B0 is the smaller baseline model. SegFormer-B1 / MiT-B1 is the larger final model. Both models are adapted to FoodSeg103 with newly initialized 104-class output layers, meaning the final prediction layer is trained for the FoodSeg103 labels instead of reused from another segmentation dataset.

These prior works provide context for the segmentation problem, but the experimental comparison in this paper is intentionally focused on the SegFormer variants. This allows the study to isolate the effect of increased model capacity, while handling visible composition separately. The paper therefore does not claim an architecture-wide comparison against FCN, DeepLabV3+, U-Net, BEiT, or published FoodSeg103 benchmark systems.

## 3 Methodology

### 3.1 Dataset and Label Handling

The FoodSeg103 split used in this study contains 4,983 images in the original training split and 2,135 images in the held-out test split. A seeded 10% validation split is allocated from the original training split, resulting in 4,485 training images and 498 validation images. The official test split is kept separate for final evaluation.

The label file contains 104 total class IDs: background is class 0, and the remaining IDs correspond to ingredient categories. The value 255 is reserved as an ignore value rather than a food class; if it appears in a mask, the training and evaluation code skips that pixel so uncertain or invalid regions are not counted as either correct or incorrect. Label reduction is disabled because FoodSeg103 already uses the intended class IDs directly. Label reduction is useful for datasets where background or void labels must be removed and the remaining classes shifted down, but using it here would incorrectly turn background into an ignored value and shift ingredient IDs away from their correct FoodSeg103 class names.

### 3.2 Models

Table 1 summarizes the two SegFormer configurations evaluated in this study. Both models use ImageNet-pretrained MiT backbones and newly initialized 104-class output layers.

**Table 1: SegFormer model configurations**

| Model | Checkpoint | Role |
|---|---|---|
| SegFormer-B0 / MiT-B0 | nvidia/mit-b0 | Smaller baseline model |
| SegFormer-B1 / MiT-B1 | nvidia/mit-b1 | Larger final model |

### 3.3 Training Setup and Preprocessing

To make the B0/B1 comparison fair, both models use the same split, input size, optimizer, learning-rate schedule, number of epochs, batch size, random seed, and checkpoint-selection rule. The only intended modeling difference is the SegFormer backbone capacity. Table 2 summarizes the shared experimental settings used for both models.

**Table 2: Shared settings for the SegFormer comparison**

| Setting | Value |
|---|---|
| Input size | 512 × 512 |
| Batch size | 8 |
| Optimizer | AdamW |
| Learning rate | 6e-5 |
| Weight decay | 0.01 |
| LR schedule | Polynomial decay |
| Epochs | 40 |
| Mixed precision | bf16 |

| Seed | 42 |
|---|---|
| Validation split | 10% of training split |
| Checkpoint selection | Best validation mean IoU |

The model requires fixed-size inputs, so images and masks are resized to the same new pixel grid 512 × 512 during training. This keeps the food regions in the image aligned with their labels in the mask. However, resizing creates new pixel locations that may fall between the original pixels, so the code must decide what value each new pixel should receive. For RGB images, bilinear interpolation is used because image pixels are color values, and averaging nearby colors produces a smoother resized image. For segmentation masks, nearest-neighbor interpolation is used because mask pixels are class IDs, not continuous values. Nearest-neighbor resizing copies the closest original class ID, preventing labels such as rice or carrot from being averaged into invalid or incorrect class numbers.

Training augmentations include random scale/crop, horizontal flip, and mild color jitter applied to images. Validation and test preprocessing are deterministic so that evaluation is consistent.

## 3.4 Evaluation Metrics

Validation mean IoU is used for checkpoint selection: after each epoch, the checkpoint with the highest validation mean IoU is saved as the best model. Final evaluation is performed separately on the held-out test split. Each test image is resized to 512 × 512 for model inference, but metrics are computed at the original ground-truth mask resolution. To do this, the model's raw output logits are resized to match the original mask size before selecting the highest-scoring class for each pixel.

The resulting predictions are accumulated into a confusion matrix and used to compute four metrics. Pixel accuracy measures the fraction of evaluated pixels assigned the correct class. Mean IoU measures the average overlap between predicted and ground-truth regions across classes, penalizing both missed pixels and extra predicted pixels. Mean IoU excluding background reports the same metric after removing the background class. Mean class accuracy averages accuracy across classes by calculating accuracy separately for each class and finally averaging those class-level accuracies. This gives rare classes more equal influence than pixel accuracy, which is dominated by large or frequent regions.

## 3.5 Visible Ingredient-area Summary

For each predicted mask, the area module counts the number of pixels assigned to each non-background ingredient class. For each predicted ingredient class (c), the visible-area percentage is computed as:

$$P_c = \frac{N_c}{N_{\text{food}}} \times 100$$

where ($N_c$) is the number of pixels predicted as ingredient class (c), and ($N_{\text{food}}$) is the total number of pixels predicted as any ingredient class, excluding background and ignore pixels.

This output is designed to be easy to interpret by answering what share of the predicted visible food pixels belong to each ingredient class. This turns a dense segmentation mask into a short, readable composition summary.

# 4 Results

## 4.1 Held-out Test Comparison

Table 3 shows the main held-out test result. Both models are evaluated on 2,135 test images, 104 labels, and original mask resolution. Figure 1 visualizes the same metrics to make the relative differences easier to compare.

**Table 3: Test results for SegFormer-B0 vs. SegFormer-B1**

| Metric | SegFormer-B0 | SegFormer-B1 | B1 – B0 |
|---|---|---|---|
| Pixel accuracy | 0.7709 | 0.7929 | +0.0220 |
| Mean IoU | 0.2521 | 0.3204 | +0.0683 |
| mIoU excl. background | 0.2457 | 0.3145 | +0.0688 |
| Mean class accuracy | 0.3590 | 0.4378 | +0.0788 |

SegFormer-B1 improves every saved test metric. The most important gain is mean IoU, which increases from 0.2521 to 0.3204, an absolute improvement of 0.0683. As shown in Figure 1, B1 also improves pixel accuracy, mean IoU excluding background, and mean class accuracy. These consistent improvements support SegFormer-B1 as the final model.

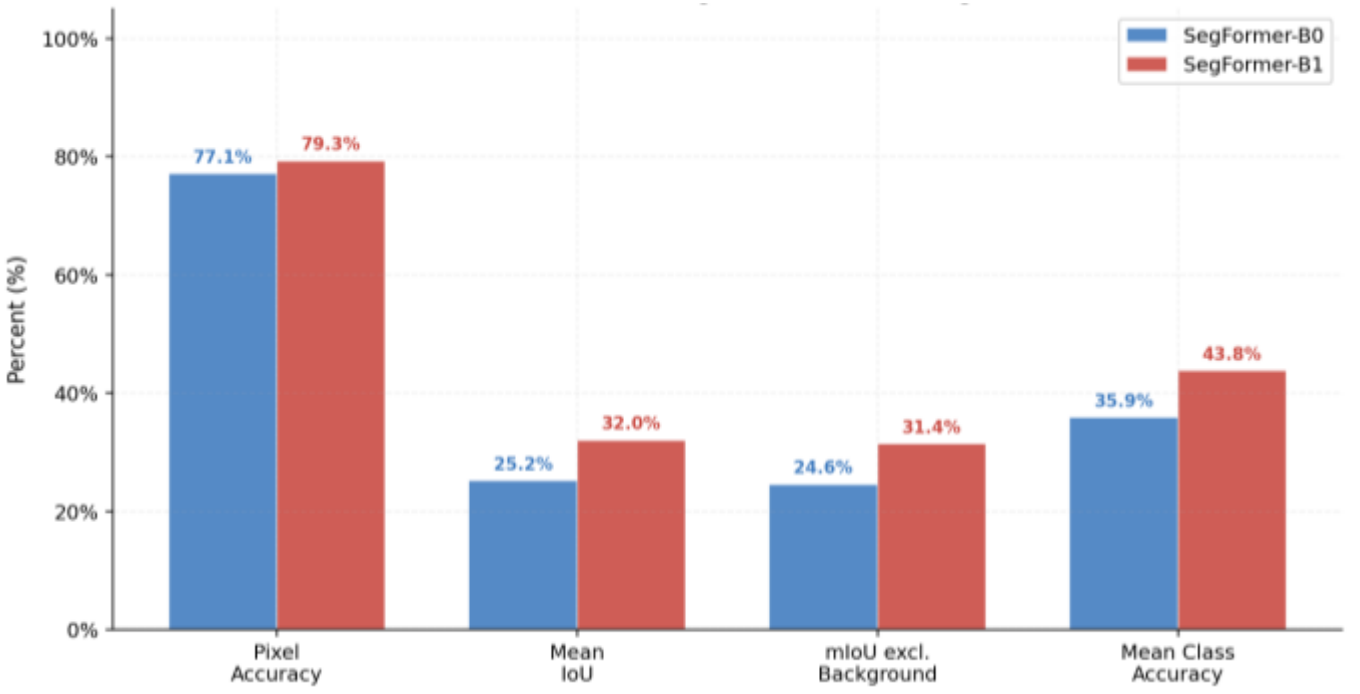

**Figure 1: Comparison of test results for SegFormer-B0 and SegFormer-B1.**

## 4.2 Validation Learning Curves

Figure 2 shows the validation mean IoU curves for SegFormer-B0 and SegFormer-B1 over 40 training epochs. SegFormer-B1 maintains higher validation mIoU than B0 throughout training and reaches its best validation mean IoU of 0.3403 at epoch 34. SegFormer-B0 reaches its best validation mean IoU of 0.2953 at epoch 37. This pattern suggests that the larger B1 backbone learns stronger validation masks under the same training settings.

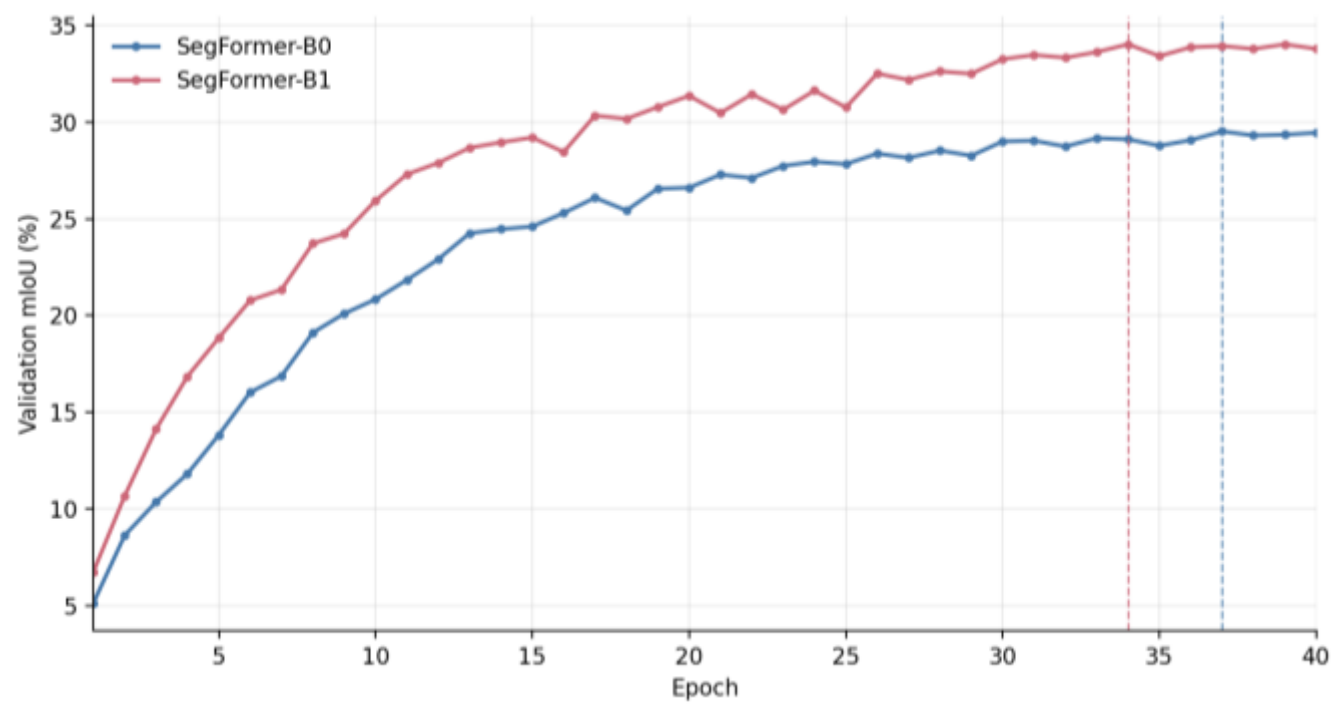

**Figure 2: Validation mean IoU over 40 training epochs.**

## 4.3 Per-class IoU Analysis

Mean IoU is much lower than pixel accuracy because FoodSeg103 is a many-class, long-tail segmentation task. Large and frequent regions can dominate pixel accuracy, while mean IoU gives each class more equal influence. This makes small or rare ingredients important to the final score.

Table 4 lists the ten strongest non-background classes for SegFormer-B1. These classes are often visually distinctive or occupy recognizable regions, which may make them easier for the model to segment.

**Table 4: Highest-IoU classes for SegFormer-B1**

| Class | SegFormer-B1 IoU |
|---|---|
| broccoli | 0.8339 |
| corn | 0.8279 |
| green beans | 0.7829 |
| shellfish | 0.7750 |
| kiwi | 0.7629 |
| carrot | 0.7498 |
| lemon | 0.7322 |
| strawberry | 0.7281 |
| rice | 0.7067 |
| banana | 0.6803 |

In contrast, the bottom ten B1 classes in the saved per-class table have IoU values of 0.0: egg tart, pudding, tea, apricot, date, peanut, cashew, dried cranberries, wonton dumplings, and hamburg. This does not prove a single cause, but it is consistent with known food-segmentation challenges such as overlap, class imbalance, inter-class similarity, intra-class variation, and mixed ingredients.

## 4.4 Qualitative Comparison

Figure 3 shows qualitative test examples comparing SegFormer-B0 and SegFormer-B1. Each example includes the original image, ground-truth mask, predicted mask, and overlay. Consistent with the quantitative results, B1 often produces masks that better match the ground truth, but both models still struggle with mixed dishes, thin regions, small ingredients, and uncertain food boundaries.

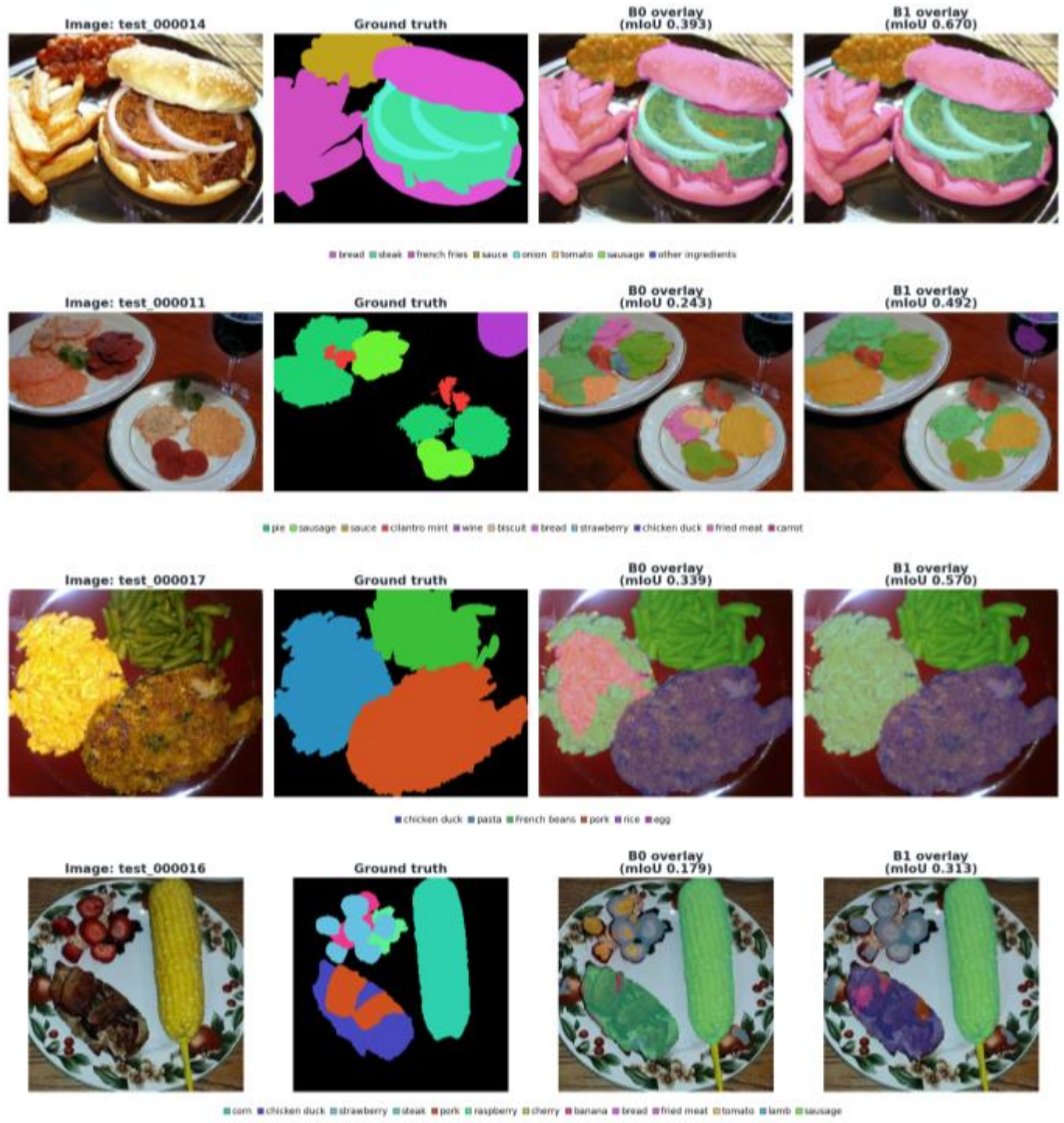

**Figure 3: Qualitative comparison of B0 and B1 predictions.**

## 4.5 Predicted Visible Ingredient Areas

Figure 4 shows an example B1 prediction converted into visible ingredient-area percentages. In this example, the predicted visible food pixels are mainly carrot, rice, and French beans, with top shares of carrot 61.8%, rice 20.0%, French beans 11.5%, chicken duck 3.1%, and shrimp 2.8%. These percentages are computed from the predicted mask and provide a visual composition summary.

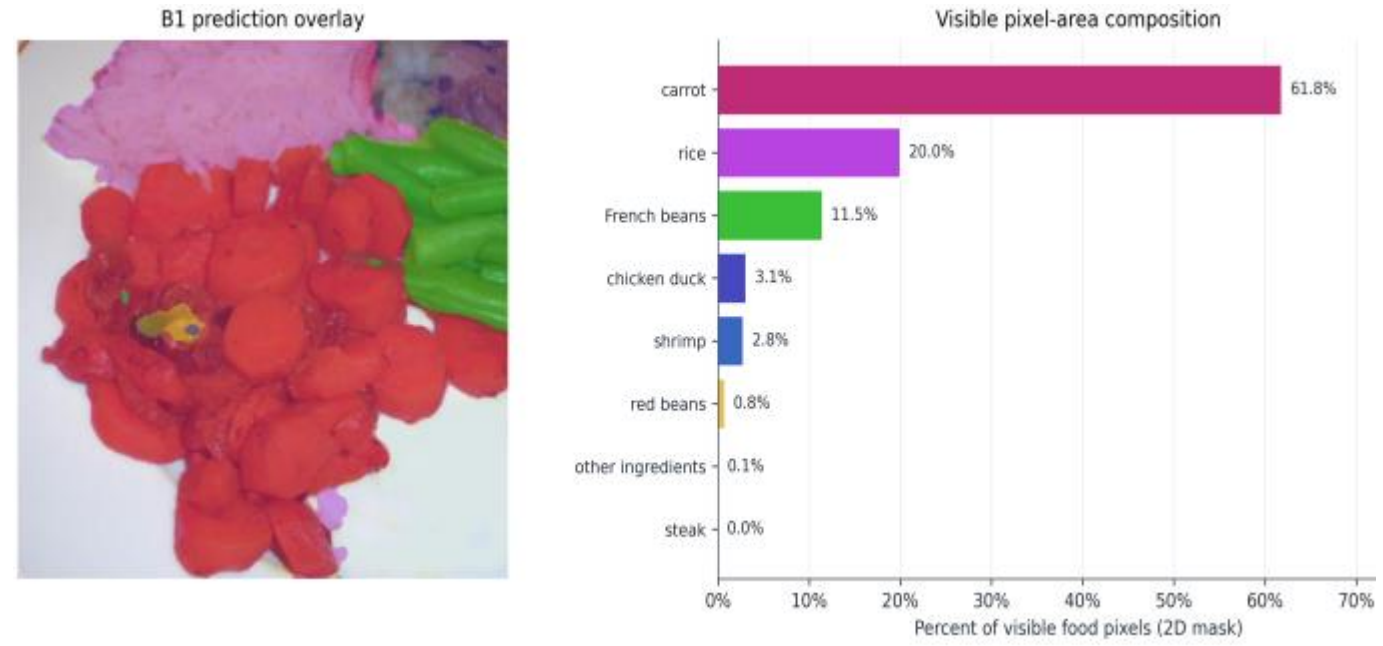

**Figure 4: B1 predicted visible ingredient-area summary.**

## 5 Discussion and Limitations

The controlled B0/B1 comparison isolates model capacity as the primary experimental change. Since the data split, preprocessing, training schedule, checkpoint selection, and evaluation code are all held fixed, the B1 improvement is reasonably attributable to the larger SegFormer variant, which learned stronger FoodSeg103 ingredient masks under this setup.

At the same time, the final mean IoU of 0.3204 shows that ingredient-level food segmentation remains difficult. This result should be interpreted in the context of prior food-segmentation research: FoodSeg103 was designed because food images need fine-grained pixel labels, and later FoodSeg103 work also identifies overlap, visual similarity, preparation differences, and class imbalance as core challenges. The observed low-IoU classes and qualitative failures match those known difficulties, especially for small, rare, mixed, or visually ambiguous ingredients.

The gap between B1's pixel accuracy and mean IoU shows why accuracy alone is not enough for this task. B1 correctly labels many individual pixels. However, its mean IoU is 0.3204, and its mean IoU excluding background is 0.3145. This suggests that foreground ingredient segmentation remains challenging even when background is excluded. Pixel accuracy is useful, but mean IoU better reflects whether the model separates each ingredient region correctly. The qualitative results support the same interpretation. B1 often captures large ingredients such as carrot, rice, beans, and fruit-like regions, but errors appear when foods overlap, when boundaries are unclear, or when visually similar ingredients appear together. These are practical failure modes for ingredient segmentation because real meals are rarely cleanly separated into simple objects.

The visible-area module has practical value because it turns a dense mask into a short, readable summary. This makes the model output easier to communicate: instead of only showing a colored mask, the system can report the predicted visible composition. The goal is not to replace calorie or macronutrient tracking with an exact automated estimate. Instead, the goal is to provide a simpler visual plate-division guidance that can reduce the need for constant weighing, measuring, or logging of ingredients. However, this feature must be interpreted conservatively. Camera angle, occlusion, plating, and ingredient thickness can all change image area without changing actual mass or nutrition. The value of the proposed summary is therefore interpretability and nutrition awareness, not dietary precision.

The main limitations are: (1) the work compares two SegFormer variants, not all major segmentation architectures; (2) FCN, DeepLabV3+, U-Net, BEiT, and other published FoodSeg103 models were not implemented as controlled baselines; (3) the results are not directly comparable to published FoodSeg103 benchmark numbers; (4) rare and small classes remain weak; and (5) the visible-area output is limited to predicted 2D image area. Future work should address class imbalance, improve small-region and boundary segmentation, and add controlled architecture baselines such as DeepLabV3+ if a broader model comparison is required. A second direction is nutrition-aware interpretation. Predicted ingredient classes could be mapped to food groups or external food-composition databases such as USDA FoodData Central [7]. This could support rough nutrition-awareness feedback, such as whether the predicted meal appears dominated by vegetables, protein-rich foods, fats, or starchy carbohydrates, which could make macronutrient tracking easier. However, exact calorie or macronutrient estimation would still require portion-size information beyond 2D image area, and image-assisted dietary assessment literature shows that fully automated dietary estimation remains challenging [8].

## 6 Conclusion

This work presents and evaluates an ingredient-level semantic segmentation pipeline for FoodSeg103. The completed experiments show that SegFormer can be fine-tuned for this task, with SegFormer-B1 selected as the final model after improving over the smaller SegFormer-B0 model under the same setup. On the test split, B1 improves pixel accuracy from 0.7709 to 0.7929 and mean IoU from 0.2521 to 0.3204.

Overall, B1 provides the strongest result in this study and performs well on several visually distinctive ingredient classes. At the same time, the remaining mean-IoU gap and low-IoU classes show that ingredient-level food segmentation is still challenging, especially for rare, small, mixed, overlapping, or visually ambiguous foods.

The visible-area summary adds value by converting predicted masks into an easy-to-read composition view. Its value is not exact dietary measurement, but interpretability: it shows how the model understands the visible composition of a meal without requiring manual ingredient-by-ingredient tracking. This makes the work a useful foundation for future nutrition-aware systems, where segmentation could later be combined with food-group mapping, external nutrition databases, or portion-size estimation. In its current form, the system is best described as an ingredient-segmentation and visible-composition system.